%% file: main.tex
\documentclass{./styles/svproc}
\usepackage{wrapfig}

\usepackage{url}

\usepackage{moreverb,url}

\usepackage{pgf}
\usepackage{tikz}
\usetikzlibrary{arrows,automata}

\usepackage[hidelinks]{hyperref}
\usepackage{algorithmic}

\usepackage{times}
\usepackage{multicol}
\usepackage[LGR,T1]{fontenc}

\usepackage{amsmath}    

\usepackage{amsthm}    
\usepackage{graphicx}   
\usepackage[font=footnotesize,labelfont=footnotesize]{caption}
\usepackage{verbatim}   
\usepackage{color}      
\usepackage{mathtools}  
\graphicspath{{./figures/}}

\usepackage{hyperref} 
\usepackage{multirow}
\usepackage{rotating}
\usepackage{array}
\usepackage{xspace}
\usepackage{indentfirst}
\usepackage{amssymb}
\usepackage{float}
\usepackage{comment}
\usepackage{sidecap}
\usepackage{enumitem}

\usepackage[ruled,vlined,noend]{algorithm2e}
\usepackage{color}
\definecolor{darkblue}{rgb}{0.15,0.09,0.3}

\SetCommentSty{mycommfont}

\definecolor{darkred}{rgb}{.3333,0.0,0.0}
\definecolor{darkgreen}{rgb}{0.09,0.51,0.3}

\SetAlFnt{\small}
\SetAlCapFnt{\normalsize}
\SetAlCapNameFnt{\normalsize}
\usepackage{algorithmic}
\algsetup{linenosize=\tiny}

\usepackage{subcaption}

\newcommand{\vgoal}{V_{\rm goal}}
\newcommand{\pterm}{V_{\rm term}}

\newcommand{\posreal}{\real^+}
\newcommand{\fref}[1]{Figure \ref{#1}}
\newcommand{\figref}[1]{Fig. \ref{#1}}
\newcommand{\clip}{{\sc Clip}\xspace}
\newcommand{\obs}[1]{\ensuremath{\textbf{#1}}}
\newcommand{\eloop}{E^{\mathrm{cycle}}}
\newcommand{\estatic}{E^{\mathrm{static}}}
\newcommand{\relatenaked}{\mathbf{C}}
\newcommand{\relate}[2]{{#1}\underset{\relatenaked}{\sim}{#2}}

\let\emptyset\varnothing

\input{./defs.tex}

\usepackage[square, numbers]{natbib}

\begin{document}
\mainmatter              
%
%
%
\title{Every Action-Based Sensor}
\titlerunning{Every Action-Based Sensor}  
%
\author{Grace McFassel \and Dylan A. Shell\vspace*{-4pt}}
\authorrunning{Grace McFassel and Dylan A. Shell}
\institute{Department of Computer Science \& Engineering,\\ Texas A\&M University, College Station TX 77843, USA\\
\email{{gracem}|{dshell}@tamu.edu}}

\maketitle              

\newcommand{\beforesectiontitle}{\vspace*{-1pt}}
\newcommand{\aftersectiontitle}{\vspace*{-1pt}}
\newcommand{\beforethrm}{}
\newcommand{\beforeproof}{}
\newcommand{\afterproof}{\vspace*{4pt}}
\newcommand{\beforedefn}{}
\newcommand{\afterdefn}{}

\begin{abstract}
\beforesectiontitle
In studying robots and planning problems, a basic question is what is the minimal information a robot must obtain to guarantee task completion. 
Erdmann's theory of action-based sensors is
a classical approach to characterizing fundamental information requirements.
That approach uses a plan to derive a type of virtual sensor which prescribes actions that make progress toward a goal. 
We show that the established theory is incomplete: the previous method for obtaining such sensors, using backchained plans, overlooks some sensors. Furthermore, there are plans, that are guaranteed to achieve goals, where the existing methods are unable to provide any action-based sensor. 
We identify the underlying feature common to all such plans. Then, we show how to produce action-based sensors even for plans where the existing treatment is inadequate, although for these cases they have no single canonical sensor. Consequently, the approach is generalized to produce sets of sensors.
Finally, we show also that this is a complete characterization of action-based sensors for planning problems and discuss how an action-based sensor translates into the traditional conception of a sensor.
\end{abstract}

\beforesectiontitle
\section{Introduction}
\label{sec:intro}
\aftersectiontitle

In his paper \textit{Understanding Action and Sensing by Designing Action-Based Sensors}\,\citep{erdmann95understanding}, Erdmann\footnote{For clarity when reading, we often refer to Erdmann by name when making reference to his theory of action-based sensors. Unless otherwise indicated, this is a reference to \cite{erdmann95understanding}.}
defines a class of abstract sensor that describes the information a sensor ought to provide a robot; his paper identifies a sort of canonical choice for such sensors.
Summarizing that classic contribution to the literature, \citet{donald95information} writes: 
\begin{quotation}
In \citep{erdmann95understanding} Erdmann demonstrates a method for synthesizing sensors from task specifications. The sensors have the property of being ``optimal'' or ``minimal'' in the sense that they convey exactly the information required for the control system to perform the task.
\end{quotation}

But, as we will show, Erdmann's treatment may overlook certain sensor choices; indeed there may be multiple sensors which are equally ``minimal'', but only some of which have been considered in the past.

Generally, analyzing the information requirements of robotic tasks has yielded fundamental scientific insights in the past (cf.~\citep{blum1978power,donald95information, okane08on}).
But making choices about sensors that are informed by information requirements is also important for practitioners, who need to balance considerations of cost, manufacturability, and reliability \cite{censi2015mathematical}.
Whether one's concern is purely theoretical or primarily practical, incompleteness is irksome.

Erdmann's sensors are defined using \emph{progress measures}, which are real-valued functions on the state space of a planning problem that indicate how movement between states leads toward a goal. 
Given such a function, for each action, one labels regions of state space where that action makes progress, forming what are called \emph{progress cones}.
These regions must be distinguished sufficiently for the robot to determine which action to execute. 
This can be realized via \emph{action-based sensors}, sensors that output actions guaranteed to make progress, which describe a subset of the progress cones containing the current state.
As an abstraction of information attainment, such sensors do not specify which environmental features or associated technologies are actually used to compute (or evaluate) these functions.
Erdmann formalizes the idea that information an agent needs is precisely and solely that which is needed to determine how to act now.

%
Action-based sensors embody the philosophy that sensors should be designed not to recognize states, only what actions must be taken to reach a goal. Utility in reaching goals is codified via progress measures and associated cones. These notions of progress are themselves computed from plans. 
The sequence goes like this: problems/tasks require plans to the solve them, plans give progress measures, measures give cones, and cones lead to sensors.
But Erdmann focuses on backchained plans, a specific subclass of plans. It would seem that there could be broader sets of progress measures than he identifies. In fact, the situation is more dire, there are plans guaranteed to solve certain problems, but for which no Erdmann-like progress measure can be produced. 


\begin{figure}[t]
    \label{fig:ex_plans}
     \centering
     \begin{subfigure}{0.31\textwidth}
     \includegraphics[scale=0.35]{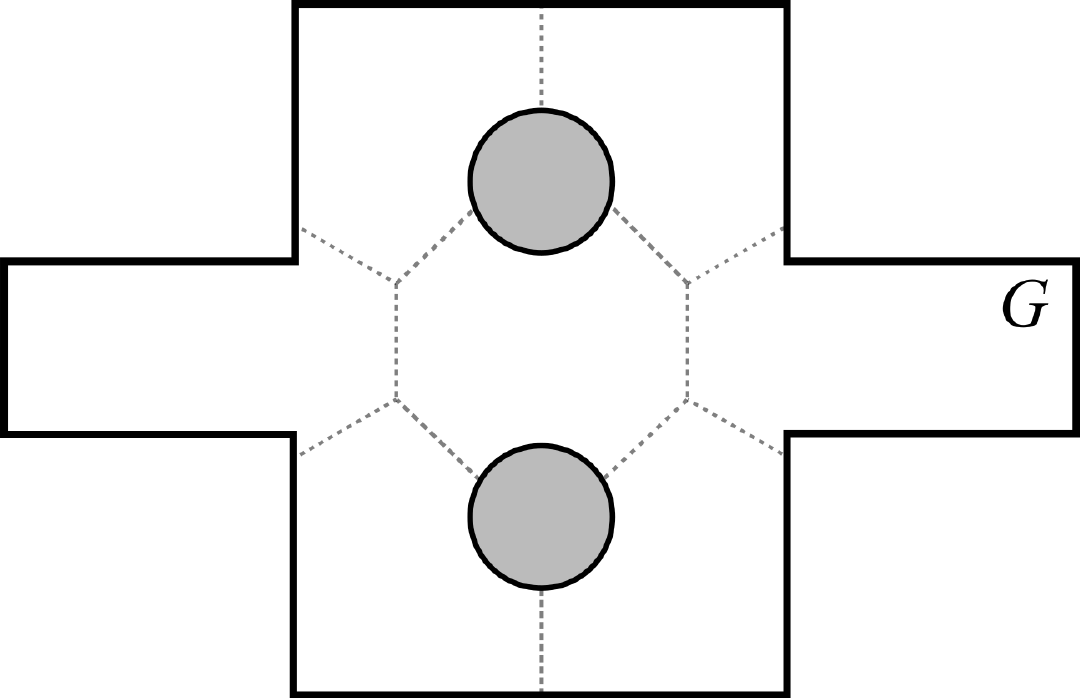}
     \caption{A basic world, with two obstacles (shaded) and a single goal region~$G$.}
     \label{fig:intro_ex_world}
     \end{subfigure}
     \hfill
     \begin{subfigure}{0.31\textwidth}
         \centering
         \includegraphics[scale=0.35]{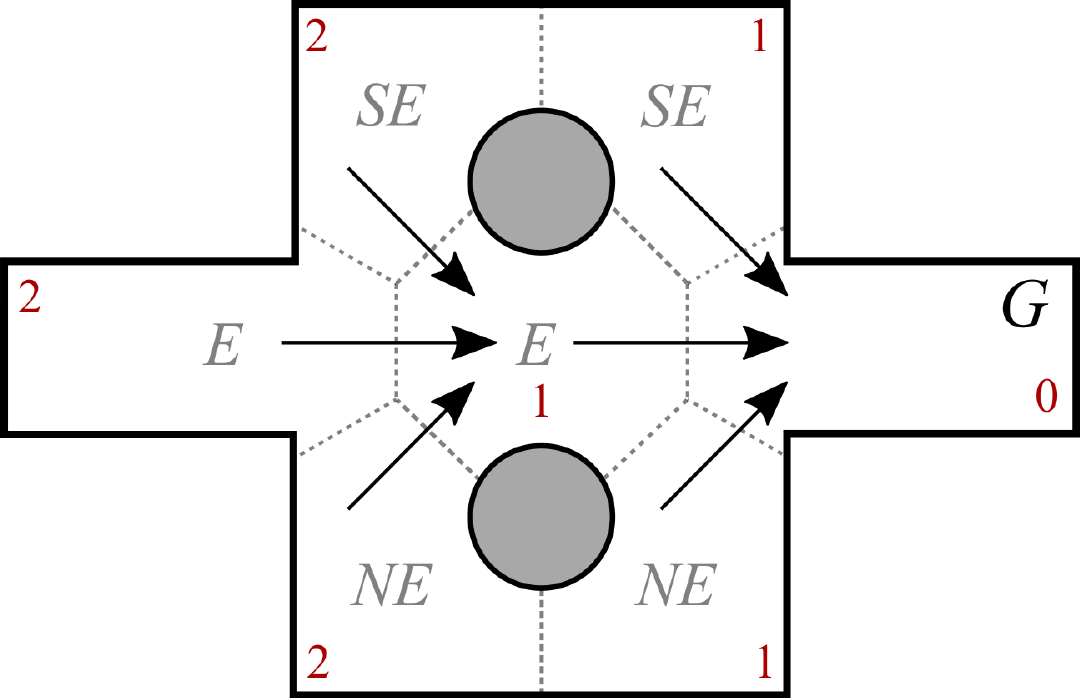}
         \caption{A plan derived via backchaining from the goal.}
         \label{fig:intro_ex_backchain}
     \end{subfigure}
     \hfill
     \begin{subfigure}{0.31\textwidth}
         \centering
         \includegraphics[scale=0.35]{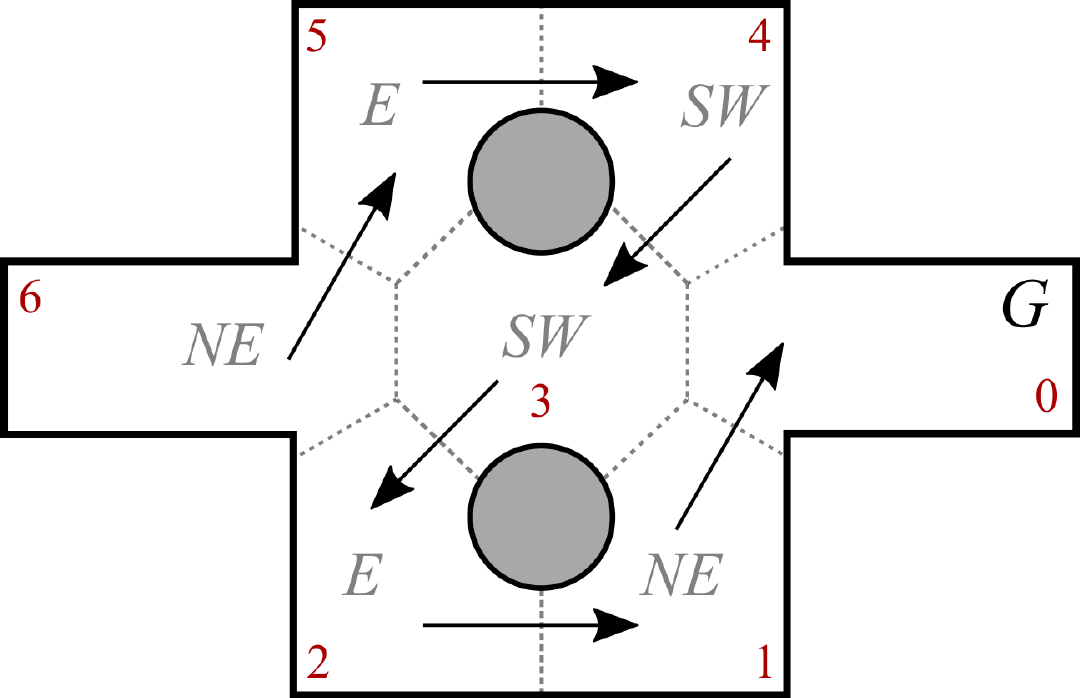}
         \caption{A slightly odder plan. {~\newline \phantom{x}}}
         \label{fig:intro_ex_splan}
     \end{subfigure}

        \vspace*{-6pt}
        \caption{A world and two different kinds of plans on that world. \figref{fig:intro_ex_backchain} uses backchaining from the goal to form a plan, while \figref{fig:intro_ex_splan} uses a Z-shaped path. This world and plans that solve it will be the basis for our discussion on the variety of plans and how this ultimately impacts sensor design. These plans can start in any region in the world, and each has a set of actions (in this case, singletons) that move them from state to state. Both plans terminate at $G$. The observations for each state are not included here, but are assumed to be distinct.
        \vspace*{-17pt}}
        \label{fig:intro_ex_2plan}
\end{figure}

Figure~\ref{fig:intro_ex_world} shows an example world. In this world, there are seven regions, each of which can be uniquely identified by the agent. Even in such a simple world, there are numerous ways an agent can get from any location to a goal, such as the two methods presented in Figure~\ref{fig:intro_ex_2plan}. We will soon see that, despite the small world and simple plans, the question of how to develop action-based sensors is not quite so straightforward!

\newpage

This paper extends the work done by Erdmann in several ways. First, we define a subset of planning problems and plans and show that within this subset, there are certain features, \emph{crossovers}, that preclude the existence of progress measures. We precisely define these features and introduce a function, \clip, that can transform a plan without a progress measure into multiple plans that do. We then show that these plans are capable of being used to define progress cones, which in turn can define action-based sensors. This obtains all action-based sensors, and we discuss how action-based sensors relate to more traditional concepts of sensing.

\vspace*{-6pt}
\beforesectiontitle
\subsection{Broader Motivation}
\label{ssec:motivations}
\aftersectiontitle

Our motivation for revisiting\footnote{Or perhaps resurrecting?} 
action-based sensors stems from an interest in what sensing, fundamentally, \textsl{is}.
Often we take sensors for granted: as a distance sensor, or wall sensor, and so on. But as \citet{brooks1993real} note: ``The data delivered by sensors are not direct descriptions of the world. They do not directly provide high level
object descriptions and their relationships.'' How we use the word ``wall''
with easy abandon! These mental categories are ingrained so deeply as to have
a pernicious influence on our thinking.  

In conceiving his theory, Erdmann asked the question of what sensors are
\textsl{for}. The action-based sensor, then,  relates what a robot should do
with what it needs to perceive. The approach conceptualizes sensors as
abstractions which entirely sidestep issues with the representation of
information to provide what is required: what action to take next.  His
definition appeared to give the utmost leeway in its requirements, being most
relaxed or unconstrained so the set of sensors seems to be maximally
inclusive\,---\,forming a sort of `free object' for sensors.  It is hardly
surprising, then, that little work has sought to expand directly upon Erdmann's
highly-original paper, for it looks to be the final word on the subject.

\vspace*{-6pt}
\beforesectiontitle
\section{Preliminaries}
\label{sec:prelims}
\aftersectiontitle
\vspace*{-6pt}

To start, two mathematical objects are introduced: planning problems and plans, where the former models tasks that entail arriving in some state in the world, while the latter prescribes actions for particular circumstances, potentially governed by internal state, to solve planning problems.  The definitions have symmetric forms reflecting how plans interleave actions with observations---actions being stipulated by the plan, observations dictated by the world's structure.

The planning problem, or world for short, is a tuple $W=(V, V_0, \vgoal, Y, U, E)$ that consists of the following:

\begin{tightitemize}
\item The finite vertex set $V$, which is the set of states in the world.
\item A set of initial states, $V_0 \subseteq V$, from where an agent may start.
\item A set of goal states, $\vgoal \subseteq V$.
\item A set of observations $Y$ which label each vertex in $V$.
\item A set of actions $U$.
\item A set of edges $E$, each of which are labeled with $\{u_0, u_1, \dots, u_k\} \subseteq U$.
\end{tightitemize}

An agent receives an observation $y\in Y$ which tells the agent where it is in the world. The agent then makes a choice of action that takes it (via an edge in $E$) from state to state, at which point it receives another observation and the process repeats. The agent begins in a state in $V_0$. Solving a planning problem demands that we develop a strategy for the agent that guarantees it will arrive in the goal region, i.e., some state in $\vgoal$, when starting from any state in $V_0$.


A plan $P = (V, V_0, \pterm, U, Y, E)$ to solve a planning problem consists of:
\begin{tightitemize}
\item The finite vertex set $V$, which is the set of states in the plan.
\item A set of initial states $V_0 \subseteq V$, from where an agent may start.
\item A set of terminating states, $\pterm \subseteq V$.
\item A set of actions $U$, subsets of which label each vertex in $V$.
\item A set of observations $Y$.
\item A set of edges $E$, which are labeled with $\{y_0, y_1, \dots, y_k\} \subseteq Y$.
\end{tightitemize}

When discussing planning problems and plans simultaneously, we will use (W) and (P) to identify to which tuple an element belongs, e.g. $V(W)$, $V_0(P)$.

\begin{figure}
\vspace*{-10pt}
    \centering
    \includegraphics[scale=0.5]{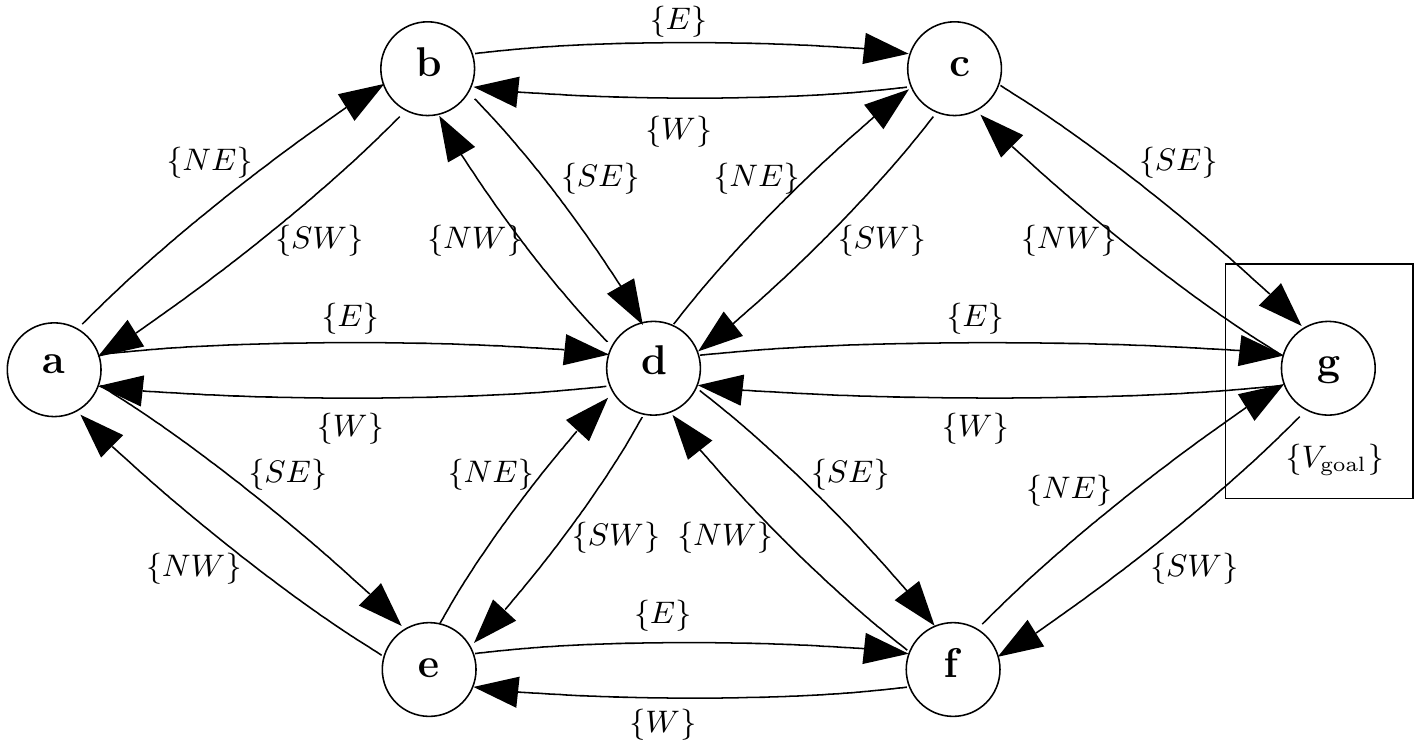}
    \caption{The example world from \figref{fig:intro_ex_world} as a connected graph. Vertices are labeled with observations \obs{a}--\obs{g}, while the actions include movement in the cardinal and intercardinal directions.}
    \label{fig:co_ex_world}
    \vspace*{-18pt}
\end{figure}

\begin{figure}[t!]
     \centering
     \hspace*{\fill}
     \begin{subfigure}[t]{0.45\textwidth}
         \centering
         \includegraphics[scale=0.35]{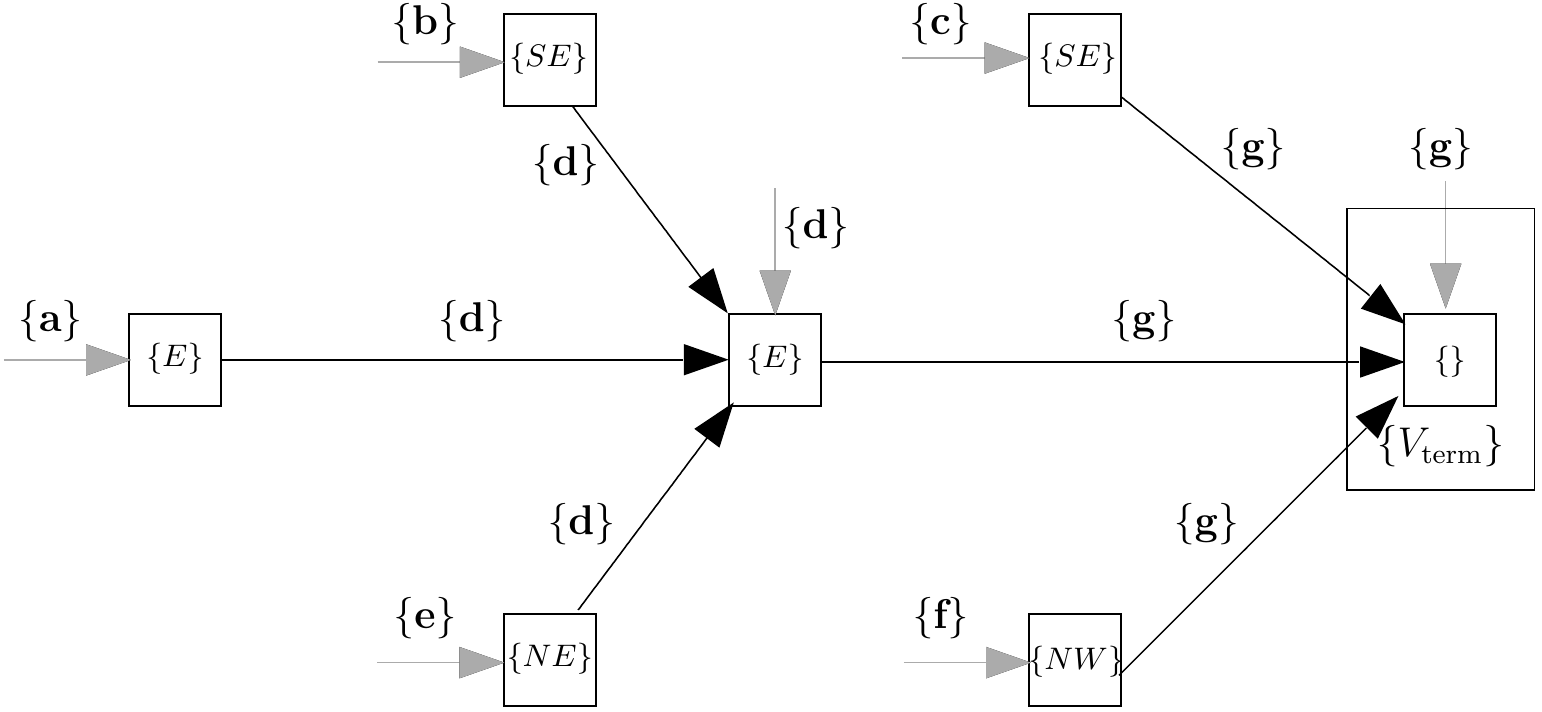}
         \caption{A plan derived via backchaining from the goal.}
         \label{fig:bc_plan}
     \end{subfigure}
     \hfill
     \begin{subfigure}[t]{0.45\textwidth}
         \centering
         \includegraphics[scale=0.35]{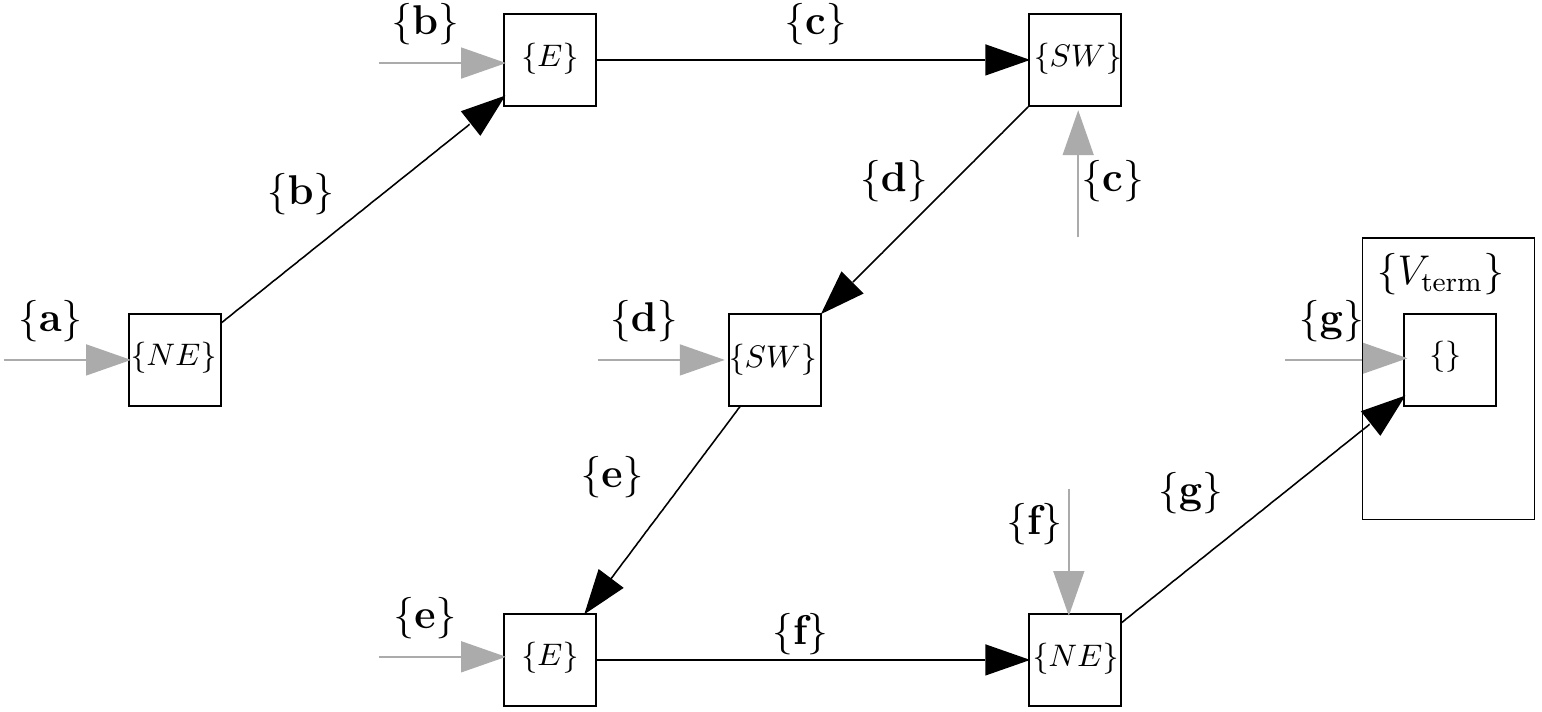}
         \caption{A slightly odder plan.}
         \label{fig:z_plan}
     \end{subfigure}
     \hspace*{\fill}
    \vspace*{-5pt}
    \caption{Our two plans from earlier, now as connected graphs. Here and in graphs that follow: lighter arrows serve as shorthand for an (elided) $V_0$ that takes no action before receiving an initial observation, whereupon it transitions to a consistent state in the plan. The lighter color rendering also serves to make the overlay of plans clearer.}
    \vspace*{-10pt}
    \label{fig:solo_zplan_graph}
\end{figure}

The preceding formalization of worlds and plans allows us to depict them as finite connected graphs.
Figures~\ref{fig:co_ex_world}
and~\ref{fig:solo_zplan_graph} show this for our earlier example. 
Plans possess terminating states $\pterm$, while worlds have goal states $\vgoal$. A vital difference is that plan edges bear observations and vertices are labeled with sets of actions; world edges carry actions and vertices have a single observation. 
An agent following a plan tracks its plan state and may take any action in the set labeling the current state. After taking an action, it receives an observation from the world, and transitions to a new vertex, itself labeled with actions, and so on. 
Every plan's initial vertices $V_0$ are labeled with the empty set, as the agent obtains a first observation to establish information about the world before taking its first action.
Only elements in $V_0$ and $\pterm$ have the empty set.

For a plan $P$ to be a solution to planning problem $W$, the plan must satisfy a few constraints: it must be \emph{safe}, in that it never attempts to take an action at a plan state if that action is unavailable at the current world state. 
Further, the plan must be ready to receive any observation that arises consistent with possible paths from the $V_0$ of the world.  An \emph{execution} is a sequence of alternating observations and actions, beginning and ending with an observation. They describe the interaction between a plan and world, giving rise to a path. Either this path must lead to states that are in $\pterm$ on the plan and in $\vgoal$ on the world, or the execution must be a prefix of a longer execution which does. If an execution is possible on both plan and world, it is a \emph{joint-execution}.
The plan must also be \emph{finite on} the world, in that all joint-executions must have bounded length.


The \emph{language} of a plan and world pair is the set of all joint-executions. 
Occasionally, we are interested in where a given execution will take us in the world. We \emph{trace} an execution $s=y_0u_1y_1u_2y_2\dots$ over the world by beginning at the vertex in $V_0$ labeled with $y_0$ and following the edges consistent with the actions in the execution. In the event of non-determinism in actions leading to multiple possible states, the following observation in the execution clarifies the resulting state in the world.

The world state reached by this method is the vertex in the world obtained by tracing a sequence $s$ on $W$, designated $\reachedv{W}{s}$.
(We will also consider tracing on a plan, and it is defined analogously.)

\vspace*{-10pt}
\beforesectiontitle
\subsection{Scope}
\label{sec:scope}
\aftersectiontitle

Consistent with Erdmann's treatment, in this paper, we examine planning problems that 
have a single goal state ($\vert \vgoal \vert = 1$) and for which $V_0 = V$, i.e.,
they may start from any state.\footnotemark 
Also like Erdmann, we begin by considering the underlying state space and then 
derive action-based sensors via a method that coarsens perceptual classes.
The result that this yields itself expresses a degree of tolerance to partial observability. Formally, this coarsening is of an original observation set, thus we begin with an 
observation set that gives direct access to the underlying states---this involves starting with planning problems that are fully observable. 

\footnotetext{In particular, knowledge of the starting location provides information in the form of context to incoming observations, allowing an agent to gather information it might not otherwise have.}

Though beyond the scope of the current treatment, 
a separate interesting problem
is to start with an original
observation set that is itself partially observable, 
perhaps to 
model some states being confounded by technological limitations, and hence
implicitly expressing information about how the world must be divided by the sensors. Since
we wish to enumerate the full set of sensors, not merely those up to some prescribed indistinguishablity, the input is a fully observable planning problem.

Some of the definitions and theory discussed throughout have overhead for handling more complicated planning problems than those just defined. 
Though we maintain definitional generality, restricting our set of planning problems simplifies discussion significantly.
We show later that this subset is sufficient to capture all action-based sensors. Relaxing these restrictions often results in problems that result in objects that are not quite action-based sensors in Erdmann's sense or which require additional interpretation to be meaningful, and are therefore beyond the scope of this paper.






\beforesectiontitle
\section{What it Means to Make Progress: the Progress Measure}
\label{sec:progress_measures}
\aftersectiontitle

To determine how plans make progress toward a goal, we start with defining what it means to make progress. Erdmann uses a framework of progress measures to develop progress cones. Given a task, to get from planning problems via progress to sensors, \citet{erdmann95understanding} prescribes: 
\begin{quotation}
\noindent\begin{itemize}[nosep]
\vspace*{-16pt}
\item[``\,1a.] Determine a sequence of actions that accomplishes the task.
\item[1b.] Define a progress measure on the state space that measures how far the task is from completion, relative to the plan just developed.
\item[1c.] For each action, compute the region in state space at which the action makes progress.\,''
\end{itemize}
\end{quotation}
\vspace*{-4pt}


The first step requires one to construct a sequence of actions, and subsequently in his paper Erdmann uses a very specific kind of plan when discussing progress measures\,---\,those obtained via backchaining from the goal. However, plans created using backchaining yield a unique progress measure corresponding to the fastest strategy which Erdmann calls ``very special''.
Our agenda is to broaden this set of plans, and doing so has implications for progress measures. In particular, we require a little more nuance, which manifests as two separate definitions in what follows.

\beforedefn
\begin{definition}[execution progress measure]\label{def:StringPM}
A \emph{progress measure over executions} on 
a solution $P$ to 
a planning problem $W$ is a function $\phi:{\pow{V(W)}\!\rightarrow \posreal}$ such that:

\begin{enumerate}[label=\alph*), topsep=0pt]
\item $\phi(V) = 0$ $\implies$ $V \subseteq \vgoal$;
\item at least one $V \subseteq \Vgoal$ satisfies $\phi(V) = 0$;
\item for any two joint-executions $p$ and $q$, if $p$ is a prefix of $q$, then $\phi(\reachedv{W}{p})>\phi(\reachedv{W}{q})$.
\end{enumerate}
\end{definition}
\afterdefn

The execution progress measure applies to sets of vertices in the world. All sets that take the value of $0$ are required to have only goals within them. There must also be at least one goal with value $0$. We restrict joint-executions, requiring that if one is a prefix of another, its value must be strictly greater. In this way, the executions of the plan visit states in the world such that the resulting progress measure is strictly decreasing.

\beforedefn
\begin{definition}[vertex progress measure]
\label{def:VertexPM}
A \emph{progress measure over vertices} on
a solution $P$ to 
a planning problem $W$ is a function $g:
V(W) \rightarrow \posreal$ such that 
$\phi_g(V)\defn \max\limits_{w\in V} \left\{g(w)\right\}$ is an execution progress measure.
\end{definition}
\afterdefn

\begin{figure}[t]
     \centering
     \hspace*{\fill}
     \begin{subfigure}[t]{0.35\textwidth}
         \centering
         \includegraphics[scale=0.35]{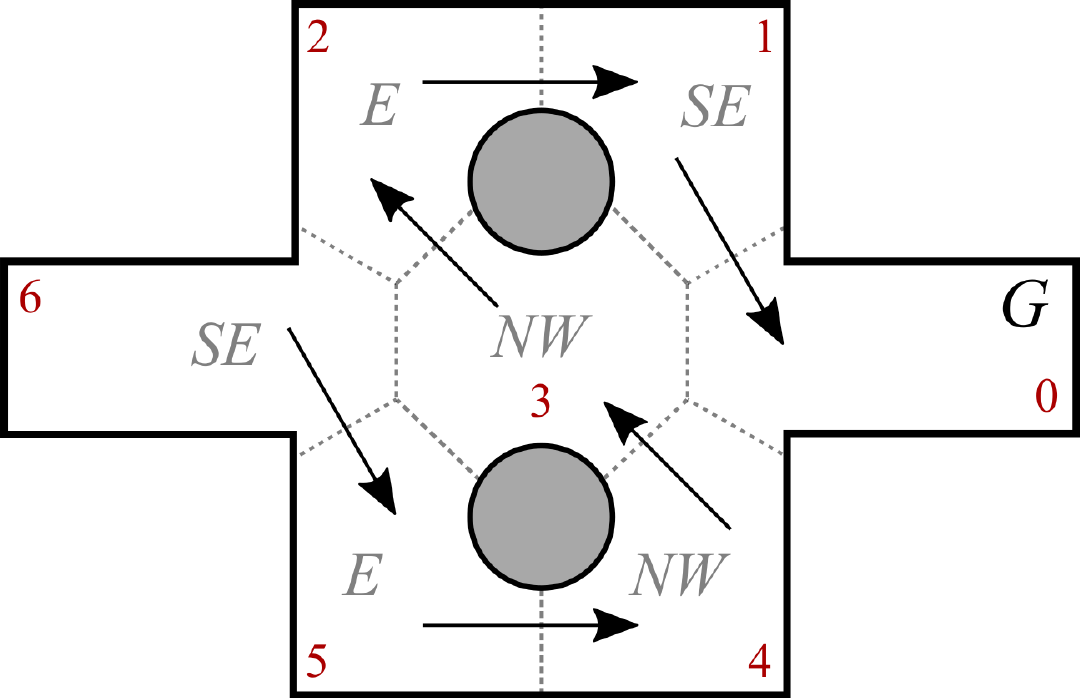}
         \caption{The mirror image of the Z-shaped plan: the S-shaped plan.}
         \label{fig:co_splan}
     \end{subfigure}
     \hfill
     \begin{subfigure}[t]{0.49\textwidth}
         \centering
         \includegraphics[scale=0.35]{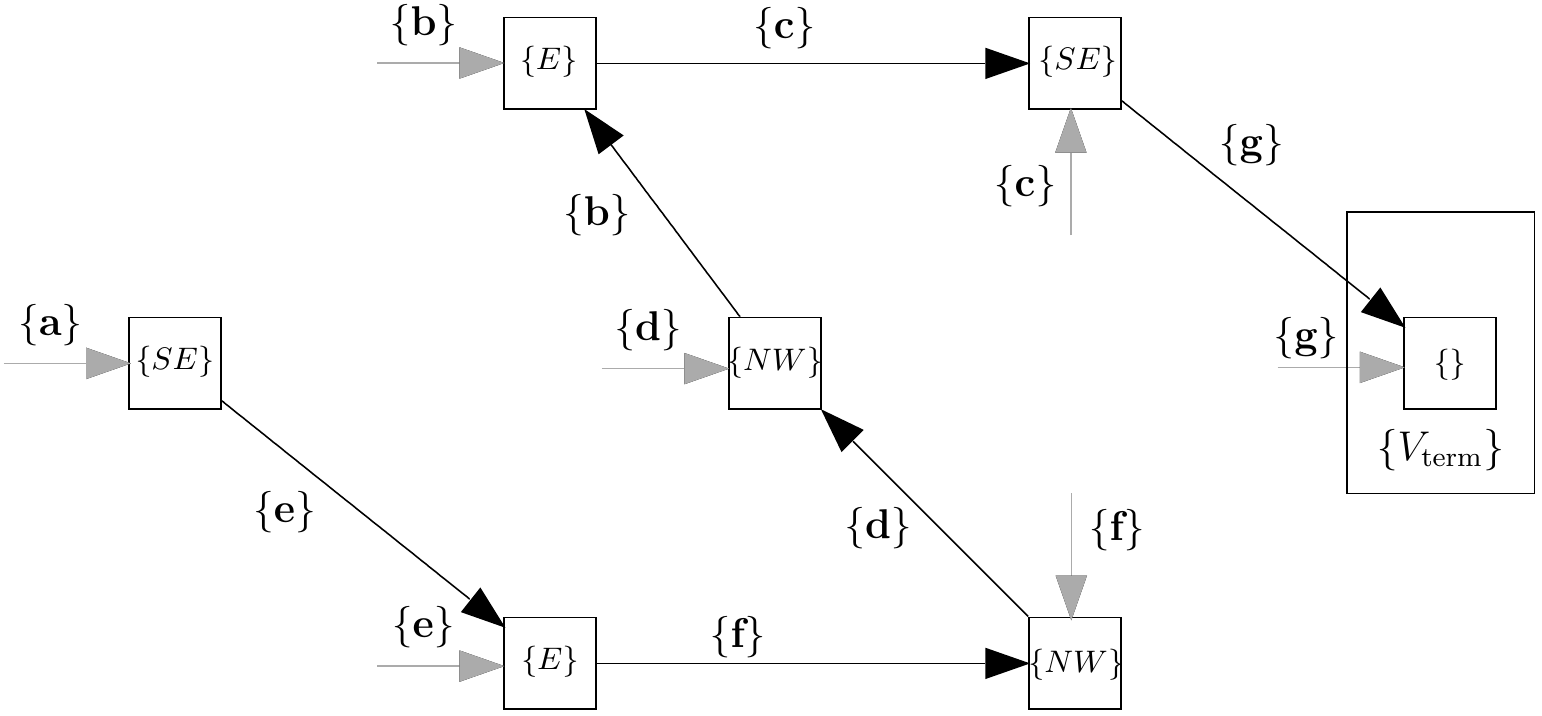}
         \caption{The new plan, as connected graph.}
         \label{fig:co_zplan}
     \end{subfigure}
     \hspace*{\fill}
        \caption{A new plan, similar to the one seen in \fref{fig:intro_ex_splan}. Although the S-shaped plan takes the same action as the Z-shaped plan at numerous states in the world, there are some states for which they are moving in opposite directions from each other.}
        \label{fig:crossover_plans}
\vspace*{-12pt}
\end{figure}

The intuition here is to give a measure on singleton states and require that we get an execution progress measure when it is lifted, in a natural way, to sets. 
For the purposes of this paper, the distinction between execution and vertex progress measures is irrelevant as we focus on small set of plans (see Section~\ref{sec:scope}) due to space limitations. 
The overhead of two separate definitions hints toward the generalization, so we have deemed to retain the distinction.\footnote{In this paper, we are considering fully observable planning problems and plans which may be at only one state in the world during any point of execution, for which an execution progress measure and vertex progress measure are equivalent. However, there exist plans in which an agent may be in multiple potential world states, for which both the execution progress measure and vertex progress measure are required to define progress-making actions.}
In light of this, we write \emph{execution progress measure} or \emph{vertex progress measure} as just \emph{progress measure}, and context will resolve any ambiguity when necessary. 

Progress measures can be seen in the red numerals in Figures~\ref{fig:intro_ex_backchain} and~\ref{fig:intro_ex_splan}. Progress measures are defined in terms of the plan's actions on the world, which can lead to a measure in which making progress leads away from the goal (in terms of distance) before reaching it. Such an example can be seen in \fref{fig:intro_ex_splan}, exemplifying the difference between a progress measure and a distance metric from the goal.

\beforesectiontitle
\subsection{Lack of Uniqueness in Progress Measures and Crossovers}
\label{sec:crossovers}
\aftersectiontitle





\begin{figure}
    \centering
    \vspace*{-10pt}
    \includegraphics[scale=0.5]{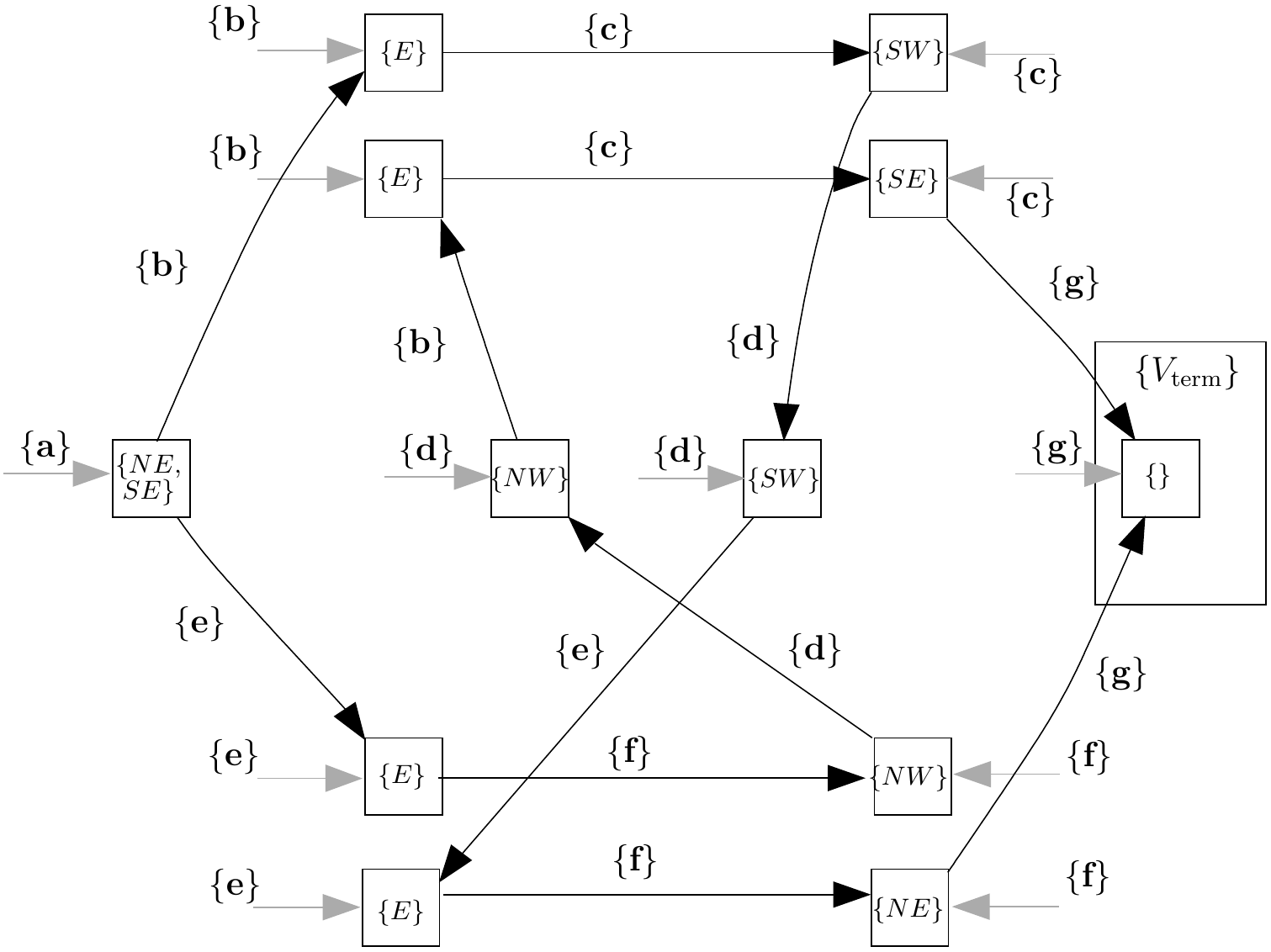}
    \caption{The S-shaped and Z-shaped plans, presented together. This plan has numerous crossovers within it. The lack of a global ordering on when states are visited is what defines a crossover.}
    \label{fig:co_szplan}
    \vspace*{-10pt}
\end{figure}

\fref{fig:crossover_plans} shows the mirror twin of a familiar plan. These two plans have some states where they take the same action, and some where they differ, creating different progress measures. There is no reason to prefer one of these plans over the other. We could even have a plan that considers both of the routes such as the plan in \fref{fig:co_szplan}, which chooses one of the two paths at random and commits to that path once it has been selected. However, an issue arises when we attempt to develop a progress measure for this plan. The issue stems from the fact that, though the plan informs our definition of progress, the progress measure is based on states in the world.

\fref{fig:co_szplan}'s plan contains executions that go in opposite directions from each other; for example, one path visits the state labeled \obs{d} before the state labeled \obs{e}, while another path visits the state \obs{e} before \obs{d}. While the plan is not structured such that the agent could actually cycle infinitely between states \obs{d} and \obs{e}, the progress measure, considering only the corresponding world states, has the impossible task of satisfying \mbox{$\phi(\obs{d}) < \phi(\obs{e}) < \phi(\obs{d})$}.

Progress measures fail to exist when there are contradictory requirements on the values that the execution progress measure must take. We refer to this issue as the \emph{crossover conflict}, or simply as the existence of \emph{crossovers}, due to the fact that the lack of progress measure extends from the fact that plan executions ``cross over'' each others' paths when traced over the world. 
In this way, crossovers can be thought of as cycles induced on the world by the plan. A plan can distinguish between multiple visits to a single world state by having multiple distinct plan states. However, the progress measure's definition considers all potential paths to and from a world state when defining values. Therefore, executions that visit states in differing orders create these cycles, resulting in a malformed progress measure.

\beforedefn
\begin{definition}[crossover conflict]\label{def:CC}
A plan $P$ on a planning problem $W$ has a \emph{crossover conflict} if it has two executions $s_1, s_2$ in the language of the plan $L_P$ such that:
\begin{enumerate}
\item $s_1$ and $s_2$ both, during their execution on $P$, visit the states $S$ and $S'$ in $W$, and
\item $s_1$ requires an execution progress measure where $\phi(S) > \phi(S')$, while $s_2$ requires an execution progress measure where $\phi(S') > \phi(S)$.
\end{enumerate}
\end{definition}
\afterdefn

To identify crossovers, the relationship between plan states, plan executions, and world states must be made explicit. One method for this is through enumeration of a plan's language and tracing that language on the world. In Section~\ref{sec:clip}, we discuss the use of a graph to identify crossovers and to avoid the need to enumerate the entire language of the plan. We are interested in identifying crossovers as their presence is the necessary and sufficient condition for the non-existence of a progress measure.

\beforethrm
\begin{theorem}
A progress measure exists iff there is no crossover within the plan.
\end{theorem}
\beforeproof
\begin{proof}
\begin{enumerate}
\item[] $\impliedby$\textbf{A Crossover Implies no Progress Measure Exists.}
The previous discussion has shown that the existence of a crossover induces an unsatisfiable condition on the progress measure.
\vspace*{3pt}
\item[] $\implies$\textbf{The Lack of a Progress Measure Implies a Crossover Exists.}
Consider a plan $P$ that solves a planning problem $W$, and which lacks a progress measure. Then, by definition of the execution progress measure, one of the following must be true:
\begin{enumerate}
\item $\phi(V) = 0, V\not\subseteq \vgoal$.
\item There is no $V \subseteq \vgoal$ where $\phi(V) = 0$.
\item There exist joint-executions $p$ and $q$ with $p$ a prefix of $q$, and $\phi(\reachedv{W}{p}) \leq \phi(\reachedv{W}{q})$.
\end{enumerate}

\medskip
Think of the progress measure not in terms of a function, but instead as an ordering on the states of the world. At least one goal must come last in the ordering, taking the value of $0$. For executions, the requirement of prefixes having a higher measure than sequences which they precede imposes an ordering on those states, as well.

If we start trying to fix the progress measure, we first assign the goal (the final element in the ordering) to value $0$ and increment up as we go earlier in the order. This resolves the issues presented by (a) and (b), should they exist.

Assume we try to correct (c) in such a way. If we are able to do so, obeying the induced ordering and assigning values to the states reached by $p$ and $q$ such that they no longer fulfill the condition of (c), then a progress measure does in fact exist. However, if it does not, then, because we were assigning values to the ordering of states based on back-tracking from the goal, we must have seen the state reached by $p$ in the ordering before the state reached by $q$ and assigned it a value accordingly. If $p$ is a prefix, that means the state reached by $p$ is visited by an execution both before \emph{and} after it visits the state reached by $q$. Therefore, there is an unsatisfiable requirement of $\phi(\reachedv{W}{p}) \leq \phi(\reachedv{W}{q}) \leq \phi(\reachedv{W}{p})$, which is a crossover.
\vspace*{-12pt}
\end{enumerate}
\vspace*{-6pt}
\end{proof}
\afterproof

Crossovers therefore are the cause of failure in a plan that does not produce a progress measure, impeding our ability to craft action-based sensors.
Recall in our example that the plan without a progress measure is created by combining two plans that do. In fact, the choice of action is what permits crossovers in the plan, but not all choices in the plan lead to crossovers. In Section~\ref{sec:clip}, we present an algorithm that identifies such choices and generates plans with progress measures when given a plan without.

\beforesectiontitle
\section{Removing Crossovers and Enumerating Progress Measures}
\label{sec:clip}
\aftersectiontitle


Given a plan $P$ that solves a planning problem $W$ and which has no progress measure, we now discuss a method to produce the set of all plans which can be derived from this initial plan, which have progress measures, and which are also solutions to the planning problem. To make precise what we mean when we say one plan is \emph{derived from} another, we define a set of actions from the world called the \emph{operative action} set.

\beforedefn
\begin{definition}[operative action set]
For a plan $P$ that solves $W$, we define the \emph{operative action set} of $P$ as a function $\mathfrak{u}_P: V(W)\rightarrow 2^{U(W)}$ such that $\mathfrak{u}_P(v)$ includes an action $u_k$ if and only if $P$ and $W$ have a joint-execution $j = y_1u_1y_2\dots u_{k-1}y_k$ for which: $j$ arrives at $v$ when traced on $W$; action $u_k$ is a label on an outgoing edge from $v$; $j$ arrives at a state labeled with $u_k$ when traced on $P$.
 \end{definition}
\afterdefn

\vspace*{-9pt}
\beforedefn
\begin{definition}[derived plan]
A plan $P'$ is \emph{derived from} another plan $P$ if the operative action set of $P'$ is contained in the operative action set of $P$, that is for all $v \in V(W)$, $\mathfrak{u}_{P'}(v) \subseteq \mathfrak{u}_P(v)$.
\end{definition}
\afterdefn
\vspace*{-6pt}

The operative action set for a plan associates each world state with a subset of that state's outgoing actions --- those that are potentially used by the plan.
Crossovers imply a cyclical dependency between world states, and therefore to identify them we need to find the correspondence between plan states and world states and clip away these cycles.
A solution is to take the entire set of operative actions and generate all possible combinations, rejecting options that aren't solutions to the planning problem or which do not have progress measures. This is a na\"\i ve approach because we don't need to consider every possible edge in the plan to remove crossovers, but only those edges that are involved in the crossover itself. For edges that are not involved in any crossover, we may take any subset of them so long as we abide by our constraints. Therefore, we propose an algorithm to generate representatives for the plan that show all possible ways of breaking crossovers without regard to the other edges.

\begin{figure}[b]
    \vspace*{-12pt}
    \centering
    \includegraphics[scale=0.5]{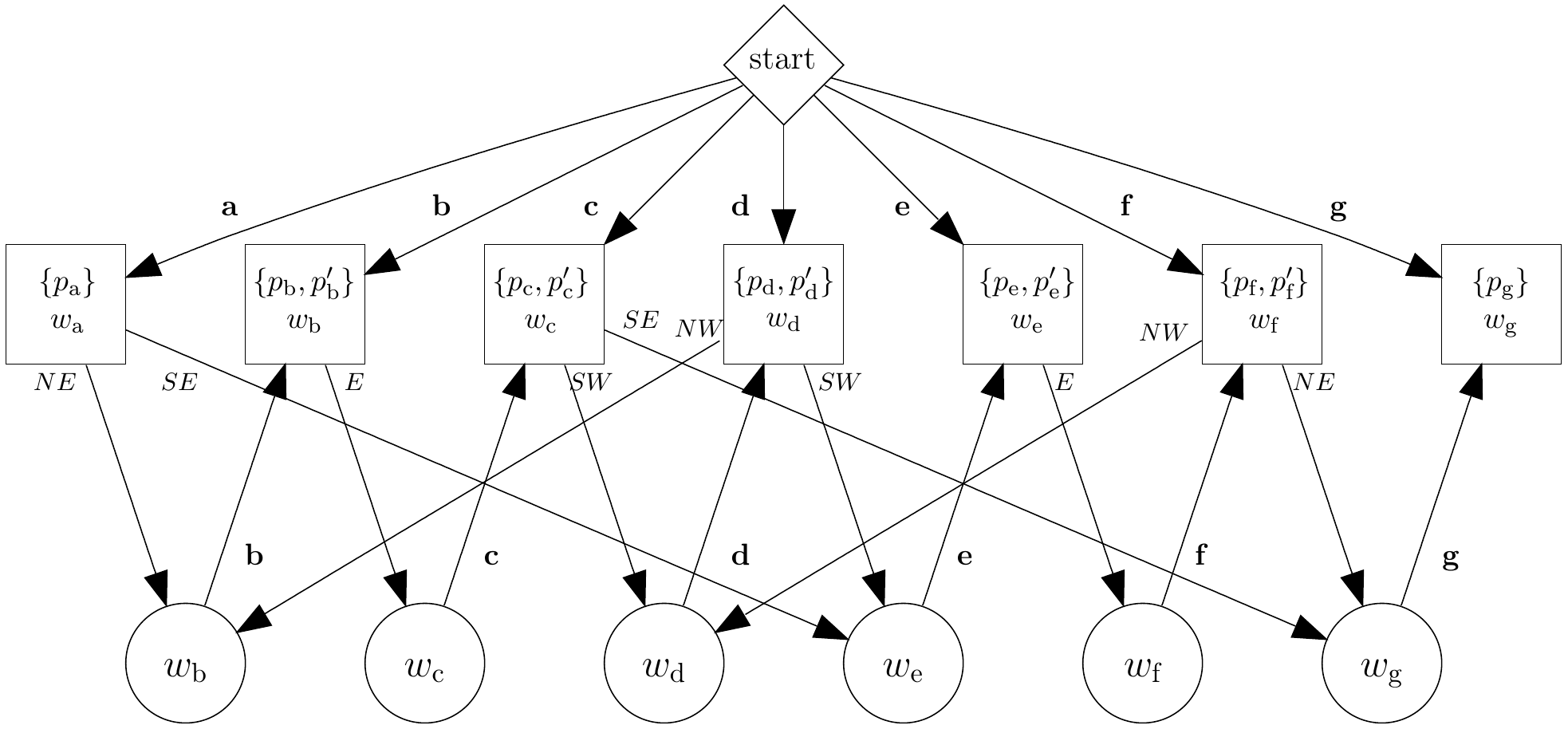}
\vspace*{-6pt}
    \caption{An I-Graph for the plan seen previously in \fref{fig:co_szplan}. Plans may have multiple vertices that correspond to the same state in the world. The I-Graph merges these into a single vertex.}
    \label{fig:igraph}
\vspace*{-6pt}
\end{figure}

To accomplish this, we present the algorithm \clip. It has two parts: first, given a plan and a planning problem, it creates a new graph structure based on the executions of the plan and the corresponding world states. We call this structure the \emph{Plan-World Interaction Graph}, or simply the \emph{I-Graph}. The I-Graph merges plan with planning problem to yield insight into the plan's movement through the world. Unlike the `flat' notion of the operative action set, the I-Graph provides this information in a more structured manner. By  finding where multiple plan states act on the same world state, it directly expresses how the plan transitions between world states.
This sequential structure helps encode the search space for edges to be removed.
In addition, because plan states that were distinct correspond to a single vertex in the I-Graph, cycles that previously existed only in terms of the progress measure now become explicit.

The I-Graph is a graph consisting of three layers: the initiating layer, the plan layer, and the world layer. The initiating layer is a single vertex. The edges of the initiating layer transition to the plan layer. They are labeled with observations that the agent may receive upon beginning execution. Each vertex in the plan layer is labeled with a pair, the first element being a subset of plan states and the second being the single world state to which they all correspond. This layer is called the plan layer because, from this layer, the outgoing edges are actions: this is the layer of the I-Graph for which the plan makes a choice. These edges go to the world layer, which corresponds to the world's choice. A single action from a state in the world may result in several different observations depending on the outcome. The world `chooses' which of these observations is obtained. The edges from this layer go back to the plan layer to a new set of plan and world states. An example appears in \fref{fig:igraph}.

{\small
\begin{algorithm}[t]
\label{I-Graph Construction}
\caption{Construction of the Interaction Graph}
\begin{algorithmic}
\REQUIRE{Inputs: World W, Plan P}
%
%
\STATE I-Graph $\leftarrow$ New Graph()
\STATE association\_queue $\leftarrow$ $\langle$P.init\_states, W.init\_states$\rangle$
\WHILE{association\_queue is not empty\:}
\STATE current\_state $\leftarrow$ association\_queue.dequeue()
\STATE plan\_states $\leftarrow$ current\_state[0]
\STATE world\_states $\leftarrow$ current\_state[1]
    \FOR{each p in plan\_states}
        \STATE new\_plan\_states $\leftarrow$ \{\}
        \STATE new\_world\_states $\leftarrow$ \{\}
        \FOR{each outgoing\_edge in p}
            \IF{outgoing\_edge has equivalent action from world\_states on world\_edge}
                \STATE new\_plan\_states.add(outgoing\_edge.target)
                \STATE new\_world\_states.add(world\_edge.target)
            \ENDIF
            \IF{new\_world\_states is a set already in the I-Graph as q}
                \STATE add edge to I-Graph: current\_state, outgoing\_edge, q
            \ELSE
                \STATE add new vertex to I-Graph: new\_world\_states
                \STATE add edge to I-Graph: current\_state, outgoing\_edge, new\_world\_states
            \ENDIF
\STATE            association\_queue.enqueue($\langle$new\_plan\_states, new\_world\_states$\rangle$)
        \ENDFOR
    \ENDFOR
\ENDWHILE
\RETURN{I-Graph}
\end{algorithmic}
\end{algorithm}
}

\beforethrm
\begin{lemma}
\label{lemma:mplan_ua}
The set of all actions for I-Graph $I$ generated from plan $P$ has the same operative action set as $P$ on the original planning problem $W$.
\end{lemma}
\beforeproof
\begin{proof}
The construction associates world states by tracing $P$'s executions.
\end{proof}
\afterproof



Our goal, and the function of the second part of \clip, is to use this I-Graph to develop plans with progress measures. We define the \emph{comes-before} relation, which indicates which world states precede each other during execution of the original plan $P$. Crossovers are now easily identified as cycles in the graph of the I-Graph, and we can both determine which world states are involved in crossovers and how many crossovers exist. For each crossover found, the cycle in the I-Graph must be clipped. However, there are many ways to clip a cycle, and cycles may even overlap with each other. We wish to enumerate all possible ways to resolve these crossovers.

\clip constructs a search tree starting from the original I-Graph. For each crossover found, \clip generates a set of candidate edges. Candidate edges are those that transition between world states in the cycle and come from a world state that has more than one outgoing action. This second requirement is because, should we remove all outgoing actions from a world state, any resulting plan cannot reach the goal from that state and therefore is not a solution.

The search tree considers the powerset of the candidate edge set for removal. For each set of edges from the powerset, we create a new child where these edges have been removed from the I-Graph and move on to the next cycle. The empty set is included in this search as cycles may share edges or states in common, and therefore the choice of removing no edges by a single cycle effectively defers the choice of which edges to remove to later.
Because \clip only removes edges, any I-Graph that has isolated a world state is invalid and is removed from consideration. To improve the search, nodes in the search tree are checked to see if they are invalid before \clip generates their children.

%
%
We now examine the output of \clip. \clip produces numerous subgraphs of the I-Graph, each of which has its own method of resolving the crossovers. We use these to define a set of plans: the \emph{representative plan} for an I-Graph is the plan which has the same operative action set as the I-Graph used to generate it.

\beforethrm
\begin{theorem}
All plans generated by \clip using a plan $P$ are derived from $P$, have progress measures, and solve the planning problem $W$.
\end{theorem}
\vspace{-\topsep}
\beforeproof
\begin{proofsketch}
\begin{enumerate}[topsep=0pt]
    \item \textbf{All plans generated by \clip are derived from the plan $P$.}
    
    \clip only removes edges from an input plan, and cannot add actions to output plans that are not part of the input. \clip creates an I-Graph, which via Lemma \ref{lemma:mplan_ua} has the same operative action set as $P$, and then generates plans from subgraphs. Therefore, any plan \clip generates must have an operative action set that is equivalent to $P$, or a subset.
    
    \item \textbf{All plans generated by \clip have a progress measure.}
    
    By the comes-before relation, \clip verifies that no cycles in the world exist before acceptance. As lack of cycles is indicative of a lack of crossovers and therefore sufficient to indicate existence of a progress measure, all plans generated by \clip have a progress measure.
    
    \item \textbf{All plans generated by \clip solve the planning problem.}
    
    By definition, before accepting any solution, \clip calculates the comes-before relation. If, for any world state, that state does not `come before' at least one goal state, \clip rejects it. \clip also rejects any plans with cycles still present. Therefore, any plan \clip accepts is a solution.
\qed
\end{enumerate}
\end{proofsketch}
\afterproof

We call the output plans of \clip the \emph{representatives} of the set of all solutions, so named because any plan in this set of desired solutions that is not generated by \clip directly can be derived from a representative itself.

\newpage

\beforethrm
\begin{theorem}
All plans derived from \clip's representatives are in the solution set.
\end{theorem}
\beforeproof
\begin{proof}
We define plans derived from \clip's results as any plans constructed from a subset of edges from a result produced by \clip. Excepting such plans that are not a solution to the planning problem, a plan with some of its edges removed will remain a solution. In addition, as removing action edges cannot induce a cycle on a plan that does not have one, the resulting plan keeps a progress measure, ensuring that it is also part of the solution set.
\end{proof}
\afterproof

\beforethrm
\begin{theorem}
Every plan in the solution set can be derived from a representative plan.
\end{theorem}
\beforeproof
\begin{proof}
Assume that there is a plan $P'$ that is not generated by \clip nor derived from \clip's solutions. Then $P'$ must solve the original planning problem $W$, use actions found only in the operative action set of $P$, have a progress measure, and not be a representative produced by \clip or derived from these results.

We define two sets: $\eloop$ and $\estatic$. $\eloop$ is the set of edges in the world that are part of the operative action set of all cycles in the I-Graph. By definition, \clip considers $\eloop$ as candidates for removal. $\estatic$ is the set of all other action edges in the operative action set. As $P'$ uses the operative action set that is also used to generate the solutions of \clip, it must differ from any given plan provided by \clip in $\estatic$ or $\eloop$ (or both).

If $P'$ differs in $\estatic$, it can be in one of two ways: either it contains an element not in any representative produced by \clip, or it is a subset of any given representative's $\estatic$. If it is a subset, then $P'$ is actually derived from that representative. If it contains an element not in a representative, then that element is not from the operative action set, and $P'$ is not truly in the solution set, as \clip does not remove any edges from $\estatic$. Therefore all representatives generated by clip contain the entire set of $\estatic$ from the original plan $P$.

If $P'$ differs from $\eloop$, it can be in one of two ways: if it has an extra edge, then it either contains an action not from the operative action set or it still contains a cycle that causes its language with the world to not to be finite. If it is smaller than any representative, than it implies \clip does not enumerate all possible values of the cycle edges. As \clip enumerates all cycle edge possibilities (through generating the powerset), it must enumerate all possible values of the cycle edges, so this is a contradiction.
\end{proof}
\afterproof

The input plan therefore yields numerous representatives, all of which have progress measures and which can be used to generate the entire set of plans of interest. 
Having resolved the issue of plans without progress measures, we next turn to the question of how to use progress measures to obtain sensors.

\beforesectiontitle
\section{Translating Progress Measures into Sensors}
\label{sec:sensors}
\aftersectiontitle

Once we have obtained a plan with a progress measure, we want to use it to link actions with how to make progress.
We achieve this through defining a \emph{progress cone}, a term inherited from Erdmann. 
Every action $u \in U$ has an associated progress cone. This progress cone is a set of observations. At any state in the world labeled with an observation in this set, the action $u$ makes progress toward the goal (transitioning from a higher-valued state to a lower-valued state) according to the progress measure. 

\beforedefn
\begin{definition}[progress cone]
\label{Def:ProgCones}
For a planning problem $W$ and plan $P$ with a  progress measure $\phi$, 
the \emph{progress cone of an action} $u \in U(W)$ 
is the largest subset of $Y(W)$, $\{y_1, y_2, \dots, y_k\}$ where
$u$ makes progress under $\phi$, from all states labeled with some $y_i$.
\end{definition}
\afterdefn

Two views of the cones are possible. The first, which is natural from the preceding definition, maps from actions to states (or, equivalently, observations); the second asks which actions make progress at a state (or given an observation).
Since both views are useful, we consider the progress cone to be a relation between observations and actions so that each observation has an associated set of actions that make progress.

\beforedefn
\begin{definition}[cone relation]
For a planning problem $W$ and plan $P$ with a  progress measure $\phi$, the \emph{cone relation} $\relatenaked \subseteq Y(W)\times U(W)$ contains $(y,u)$ if there exists a progress cone for the action $u$ containing $y$.
We will also write $\relate{y}{u}$, when
$(y,u) \in \relatenaked$.
\end{definition}
\afterdefn

As the planning problem must be solved from any location in the world, any observation $y$ that is used as a label of a state in $V(W)$ must have at least one action $u$ where $\relate{y}{u}$. This forms a covering over the set of observations $Y(W)$.

For an observation, there are potentially many progress-making actions. However, only one action is needed for any given $y$ to guarantee that the agent will eventually arrive at the goal. 
We can define a class of functions that, for each observation $y$, return a single action $u$, transforming the covering into a collection of partitions. We call these functions \emph{singleton action-based sensors}.

\beforedefn
\begin{definition}[singleton action-based sensor]
\label{def:single_actionsensor}
A function $f: Y(W) \rightarrow U(W)$ is a \emph{singleton action-based sensor} if $\relate{y}{f(y)}$
 for every $y$.
\end{definition}
\afterdefn


The connection between singleton action-based sensors and real sensors is not immediately apparent. A ``traditional'' sensor can be represented as a function $s: V(W) \rightarrow Y(W)$, taking world states as inputs and returning some observation.

To bridge the gap we define a new set of observations $Y' = \{ y'_u \mid u \in U(W)\}$ by making a correspondence of each element to an action, as indicated by the subscript.
For an element $y\in Y(W)$, $y \mapsto y'_u$ if $f(y) = u$ according to a singleton action-based sensor.
Recall that in our framework, each element in $Y$ mapped to a single state in the world, so we can think of the elements of $Y$ as a kind of stand-in for the world states of $W$. $W \rightarrow Y \rightarrow Y'$ is therefore equivalent to $W \rightarrow Y'$, and takes in states in the world and maps them to this new set.




The process above transformed a covering into a partition through use of a function. However, perhaps we would like multiple (or even all) possible progress-making actions for a single observation. To achieve this, we define \emph{permissive action-based sensors}.


\beforedefn
\begin{definition}[permissive action-based sensor]
\label{def:perm_actionsensor}
A function $f: Y \rightarrow 2^U \setminus \{\emptyset\}$ is a \emph{permissive action-based sensor} if, for every $u \in f(y)$, $\relate{y}{u}$.
\end{definition}
\afterdefn

The relationship of this object to more traditional sensors is less clear than for a singleton action-based sensor. We can construct a $Y'$ as before, where now each element $y'$ corresponds to some subset of $U$, but the semantics of actions within multiple different sets becomes a matter of interpretation.

\begin{wrapfigure}[17]{r}{0.5\textwidth}
    \vspace*{-26pt}
    \centering
    \includegraphics[width=0.48\textwidth]{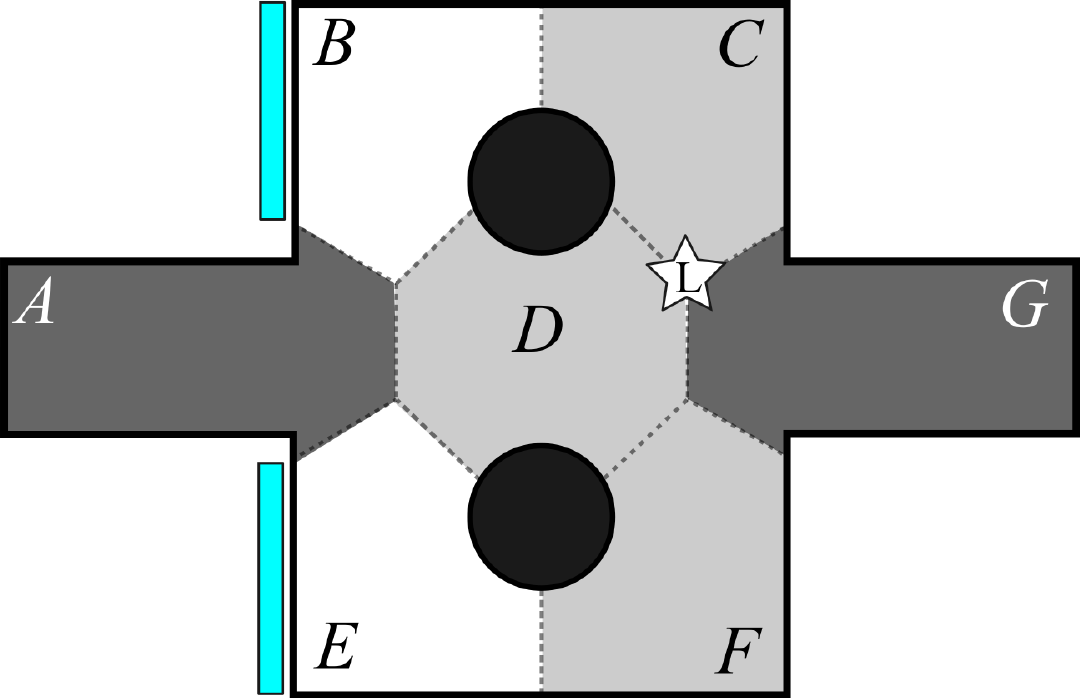}
    \caption{A world in which we may have confusion in states that precludes Erdmann's sensors being functional. States \textbf{C} and \textbf{D} both are in mid-range lighting and can see the landmark $L$, making them indistinguishable. For such a sensing setup, the partition required by the backchained plan cannot be realized.}
    \label{fig:odd_sensors}
\end{wrapfigure}

\beforesectiontitle
\section{Example}
\label{sec:example}
\aftersectiontitle

By defining action-based sensors for a larger family of plans, \clip produces novel sensor options compared to those obtained by backchaining. Additional outputs can be important in practical implementations where factors like
sensing technology and cost must be considered.
We illustrate this by adding to the running example. \fref{fig:odd_sensors} shows our example environment in a bit more detail. Windows (in blue) allow light to enter and result in varying light levels throughout the space. A landmark (L) can be seen from states $C$, $D$, and $G$, but not elsewhere.
Consider a scenario in which a robot can only recognize brightness levels and the presence of a nearby landmark. Such a robot cannot distinguish states $C$ and $D$. This presents a problem for the action-based sensor obtained from a backchained plan such as that in \fref{fig:intro_ex_backchain}, which requires these states be distinguished. The plan in \fref{fig:intro_ex_splan}, however, takes the same actions in states $C$ and $D$, and therefore describes a partition of the world that can be realized with this setup. The action-based sensors defined by the extended family of plans we consider allow one to better
respect the constraints that designers may actually face.


\beforesectiontitle
\section{Completeness}
\label{sec:completness}
\aftersectiontitle

We have presented the limitations of the existing theory of action-based sensors and
extended it to cover a broader group of plans. As this paper attempts to fill a
gap in Erdmann's treatment of action-based sensors, there is the question of
whether the proposed method is itself comprehensive. 
We claim that the method presented here captures all action-based sensors obtainable for any
given plan $P$ that solves a planning problem $W$ satisfying the
criteria in Section~\ref{sec:scope}.

\noindent\textbf{We Capture All Action-Based Sensors.}~\emph{Proof:}
Assume there is some action-based sensor that is not obtained by the method described above. 
This action-based sensor maps from the set of observations $Y$ to sets of actions $2^U \setminus \{\emptyset\}$. 
This sensor must be usable to solve the planning problem. Therefore, this sensor must give an action that proceeds toward the goal from each state in the world. This induces an ordering on the states, which means that a progress measure can be created on the world using this sensor. As this progress measure gives an action from each state in the world, it prescribes a plan that is capable of starting from any location. But then it is included in the set of plans that we consider. Therefore, this action-based sensor is obtainable.
\qed

\beforesectiontitle
\section{Conclusion}
\label{sec:implications}
\aftersectiontitle

Having extended the work done by Erdmann we can now obtain not only the sensors that correspond to the backchained solution, but the complete set of all action-based sensors. Given a library of components, this set of sensors can guide practitioners in determining what is simultaneously realizable, respectful of environmental constraints, and sufficiently powerful to accomplish the task. Design and selection of sensors then becomes just a question of intersecting constraints.

One caveat to this notion of completeness is that progress measures, by their nature, do not make use
of state. Our example plan with crossovers chooses a path at the start of its execution, a choice that is encoded within the plan structure itself through additional vertices.
Such additional structure is required for this plan to succeed. Rather than
thinking of crossovers as simply cycles to untangle, they imply that execution
of a given plan requires some internal memory in order to make progress toward
the goal.
The idea of sensors that memorize information in order to give relevant actions goes well beyond what we typically consider when discussing a sensor. The authors are interested in pursuing how to define and make use of these ``stateful'' sensors for more complex behavior. This hints that adopting a sensor-based perspective may give new ways to understand plans.

\vspace*{-8pt}
\subsection*{Acknowledgement}
{\small
\vspace*{-6pt}

This work was supported, in part, by the National Science Foundation through awards IIS-1453652 and IIS-1849249, and from a graduate fellowship provided to Texas A\&M University by the 3M~Company and 3M~Gives.
}

%
%
{
\renewcommand{\clearpage}{}
\renewcommand{\bibname}{References\vspace*{-35pt}}
\bibliographystyle{plainnat}
\vspace*{5pt}
\bibliography{refs}
}
\end{document}

%% file: defs.tex
\usepackage{graphicx} 

\newcommand{\defn}{\coloneqq}

\renewcommand{\emptyset}{\varnothing}

\newcommand{\Vgoal}{V_{\rm goal}}

\def\markatright#1{\leavevmode\unskip\nobreak\quad\hspace*{\fill}{#1}}

\newcommand{\gobble}[1]{}

\newcommand{\pow}[1]{\ensuremath{{2}^{#1}}\xspace}
\newcommand{\real}{\ensuremath{\mathbb R}\xspace}

\newcounter{tecounter}
\newenvironment{tightitemize}
{
    \begin{list}{$\bullet$}{%
        \setlength{\leftmargin}{21pt}
        \setlength{\topsep}{6pt}
        \setlength{\partopsep}{0pt}
        \setlength{\itemsep}{2pt}}
    \ignorespaces}
{\unskip\end{list}}

\renewenvironment{proof}{\par\emph{Proof:}}{\markatright{$\square$}\par}
\newenvironment{proofsketch}{\par\emph{Proof:}}{\markatright{}\par}

\providecommand{\reachedv}[2]{\ensuremath{\mathcal{V}^{#1}_{#2}}}